\documentclass{article}
\usepackage[a4paper, margin=1in, top=0.5in]{geometry}
\usepackage{times}
\usepackage{graphicx} 
\usepackage{epstopdf}
\usepackage{subfigure} 
\usepackage[boxed]{algorithm}
\usepackage{algorithmic}
\usepackage{framed}
\usepackage{color}
\usepackage{mathtools}

\DeclarePairedDelimiterX{\infdivx}[2]{(}{)}{%
  #1\;\delimsize\|\;#2%
}

\usepackage{amssymb, amsmath, amsthm}
\usepackage{array}
\usepackage{stackengine}

\newtheorem{theorem}{Theorem}[section]

\newtheorem{remark}[theorem]{Remark}

\begin{document}

\title{Convolutional Normalization}

\author{Massimiliano Esposito \and
Nader Ganaba}

\date{}

\maketitle
 
\begin{abstract}
As the deep neural networks are being applied to complex tasks, the size of the networks and architecture increases and their topology becomes more complicated too. At the same time, training becomes slow and at some instances inefficient.  This motivated the introduction of various normalization techniques such as Batch Normalization and Layer Normalization. The aforementioned normalization methods use arithmetic operations to compute an approximation statistics (mainly the first and second moments) of the layer's data and use it to normalize it. The aforementioned methods use plain Monte Carlo method to approximate the statistics and such method fails when approximating the statistics whose distribution is complex. 
Here, we propose an approach that uses weighted sum, implemented using depth-wise convolutional neural networks, to not only approximate the statistics, but to learn the coefficients of the sum.
\end{abstract}
\tableofcontents

\section{Introduction}
During the past few years, there has been considerable success of applying deep learning to complex problems ranging from speech synthesis (cite WaveNet) to deep reinforcement learning achieving victories in complex games such as Go \cite{silver2016mastering}. This, in turn, has fueled massive interest in the field. At the core of deep learning are training algorithms which allow neural networks to learn from the data they are presented or learn how to extremize a target quantity. One such algorithm is the stochastic gradient descent (SGD) (cite) and it is one of most widely used training algorithms.
\\

However, this does not come without issues as many training algorithms suffer from being slow at converging to the optimal state and lack stability when approaching the optimum (citation needed).One common approach is to transform the data into a manageable form through centring and scaling the data and it is the base of many normalization techniques. In fact, the success of ``Batch Normalization'' by Sgzedy and Ioffe \cite{batch_norm} sparked an interest in such techniques. In \cite{batch_norm}, they noticed that the learning procedure, which uses stochastic gradient descent (SGD), or derived adaptive algorithms like Adam \cite{DBLP:journals/corr/KingmaB14}, can be hindered and they claimed that a phenomenon known as internal covariance shift is the cause. This phenomenon depends on the change of the network's weights during backpropagation that can lead to the shift of data-distribution towards the saturation regime of the activation functions. Although the authors of \cite{NEURIPS2018_905056c1} claim otherwise and that internal covariance shift does not cause slow convergence. In any case, to avoid this pathology, in \cite{batch_norm} they proposed re-scaling of the data in a channel-wise fashion. The increase in performance motivated research into normalizations techniques, which lead to the introduction of methods such as \textit{Layer Normalization} \cite{ba2016layer}, \textit{Instance Normalization} \cite{ulyanov2017improved} to name a few. In this paper, we present a new type of normalization procedures that apply, specifically, to images and directly generalise Batch Normalization.
\\

The standard normalization techniques rely on an arithmetic method, which are in essence is a bare basic Monte Carlo method, to compute the statistics from data and then uses it to transform the tensor, and in our case the image. That poses an issue, mainly, it can output statistics that are fictitious. In addition to that, the aforementioned normalization methods do not tackle data decorrelation, and that is because it does not approximate the covariance matrix. Indeed, naive application of Monte Carlo methods can fail when the data belong to a probability distribution that is difficult to sample from. In order to tackle such issue, we propose to use neural networks to render normalization methods more flexible and general. Further, using arithmetic methods do not take into consideration the hidden and underlying structure of the images, whereas using neural networks can uncover such structures and adjust the normalization accordingly.
\\

That said, there are many ways in which we can introduce neural networks and here we present one where a neural network is used to implement a weighted sum. It is discussed in section \ref{sec:weighted}, the neural network takes the data as the input and then outputs a weighted sum and contrary to arithmetic weighted sums, the neural network learns the coefficients of the weighted sum. In other words, the network learns from the data what the weighted sum should be to accelerate the learning process. It is similar to importance sampling \cite{kahn1950random, kahn1950random_2}, where it acts a change of measure to better sample from the intractable distributions. However, one issue with this approach is that we do not have great control on the number of parameters used. In addition to that, the quality of the approximation of the mean and standard deviation depends on the number of elements in the training dataset and, in some applications, the availability of a large dataset for training is not possible.  
\\

One possible solution is presented in section \ref{sec:inference_nn}, where it is inspired by variational methods for Bayesian inference \cite{beal2003variational},  two neural networks approximate the statistics of the hidden layers activation. In other words, the networks learn the mean and standard deviation. The parametrisation of statistics using neural networks is a widely used in application of variational Bayesian methods and, to the best of our knowledge, here it is the first time applied for normalization purposes. 
\\

The advantage of this approach is that it can approximate statistics from complex and multimodel distributions and it gives freedom in choosing a larger variety of architectures. This is the case where in the section \ref{sec:inference_nn} the numerical experiments show that networks with a fraction of the number of parameters compared to Batch Normalization achieves acceleration that, in some cases, can rival and exceed it. Furthermore, as the learning is done through solving an optimisation problem and this approach is used for principal component analysis applied on manifolds \cite{fletcher2004principal} known as principal geodesic analysis. Thus, it has the potential to learn statistics from data that belong to a non-Gaussian distribution and also data defined on non-Euclidean manifolds. This is one of the reasons why such approach is appealing, however, a study of cases of data on non-Euclidean manifolds is currently beyond the scope of this paper. 
\\

\textbf{Main contributions:}
\begin{itemize}
    \item Implement weighted sum using depth-wise convolutional neural networks for computing the mean and standard deviation. This allows for approximating the statistics from data whose underlying distribution is too difficult to sample from. Depth-wise convolutional neural networks are specifically used to perform normalization for every feature map seperately. 
    \item Introduce normalization method, where the mean and standard deviation are two neural networks. Despite its infancy, we show that this approach can accelerate converge of training of classification problem by introducing less learnable parameters than the standard normalization methods. 
\end{itemize}
\section{Background}\label{sec:background}
As we are primarily dealing with images, the data structure for a single image is the so-called $(0,3)$ tensor. It can be thought of as a stack of matrices and each matrix corresponds to a different channel. When we have a set of images, this enlarges to $(0,4)$ tensor. As using tensor notation can be taxing, a convention we follow here is that we denote tensors by variables such as $x$ for the sake of simplification and space of images by $I$.  
The set of images is called a batch, denoted by $\mathcal{B}$, and it is a subset of the input component of training set $\mathcal{D}$. The batch can be defined as $\mathcal{B} := \{ x^{(i_1)},  x^{(i_1)} \in I | i_1 = 0, \dots, N_b-1\}$, where $(i_1)$ in parenthesis indicates element of the batch with size $N_b$. 
\\

A neural network is described abstractly as
\begin{align}\label{eq:abstract_net}
x^{l} = f\left( x^{l-1}, \theta \right), \quad l=1, \dots N_l,
\end{align}
where $l$ denotes the index of the layer and $\theta$ denotes the network's parameters. In this article, we consider only convolutional neural networks, and the major component of this type of networks is the kernel, denoted by $K$  and whose size is smaller than the input image. The way it works is performs a weighted sum over the image, and the coefficients of the this weighted sum are the elements of the kernel $K$. We can have multiple kernels where each kernel extracts certain information from the image. A simple example is when the kernel is set to 
\begin{align}\label{eq:kernel}
    K = \begin{bmatrix}
    -1 & -1 & -1 \\
    -1 & 8 & -1 \\
     -1 & -1 & -1
    \end{bmatrix},
\end{align}
then the extracted information are the edges found in the image. Generally, we do not set the kernels beforehand, we choose arbitrary values and then through training we update these values so that the kernel would give useful information.

Consider the training set $\mathcal{D} := \{ (x_{i_1}, y_{i_1}) \, | \, x_{i_1} \in I, \, y_{i_1} \in \mathcal{C}  \}$, where $\mathcal{C}$ is the space of classes. The training problem can then be stated as finding the parameters $\theta$ that would minimise the cost functional
\begin{align}
S(\theta) = \sum_{i_1} L\left( f^{N_l}(x_{i_1}, \theta), y_{i_1}  \right),
\end{align}
where $L$ is the cost function and it is a map $L: I \to \mathbb{R}$. There are many choices for $L$ and for linear regression it is chosen as the $L^2$ norm. For classification problems, cross entropy is the preferred choice. To solve this minimisation problem, optimisation algorithms such as gradient descent are used. However, such methods can be inefficient and slow to converge and that can be due to the magnitude of the gradient of the loss function with respect to the network's parameters. One way to remedy this is to introduce transformation that renders the data such the gradient would not have large values that hamper the convergence rate. 
\\

\begin{figure}
\begin{center}
\includegraphics[width=0.66\textwidth]{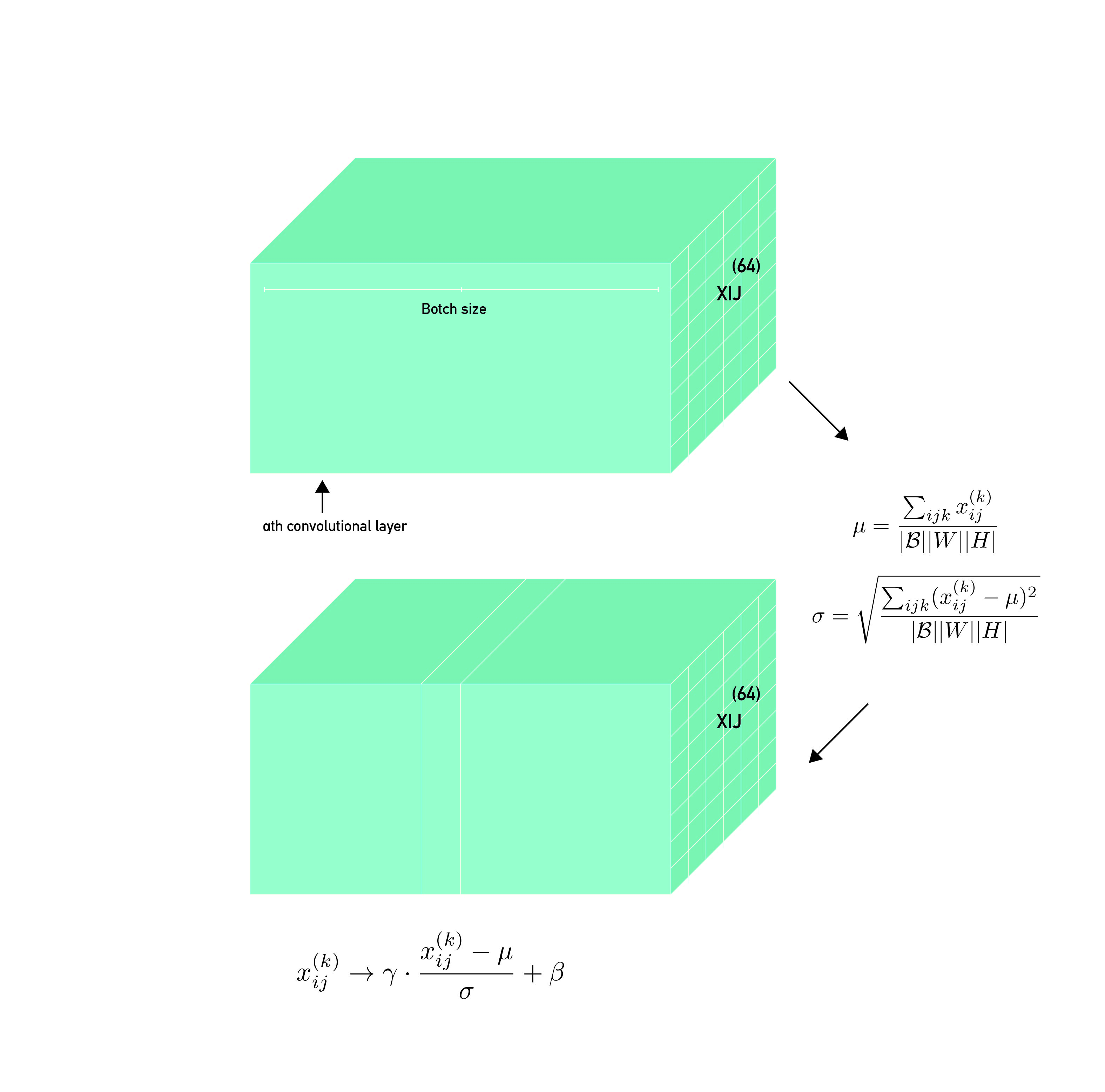}

\caption{How Batch Norm operates on images.}
\label{fig:CB}
\end{center}
\end{figure}

The most celebrated of these methods is the Batch Normalization \cite{batch_norm} and it works by centering the data to zero and scaling it so that the standard deviation becomes unity. 
Since this article's focus is on images, we recall how specifically Batch Normalization operates on them. In a deep neural network for image classification, Batch Normalization  operates per convolutional layer re-scaling each element $x_{ij}^{(k)}$ in the convolutional filter $\alpha$ as depitcted in the image below \ref{fig:CB}. Here $i,j$ denote the position of the element with respect to height and width, while $k$ denotes the $k^{th}$ element of batch $\mathcal{B}$, as we have already said.

The element $x_{ij}^{(k)}$ of the $\alpha^{th}$ convolutional layer is rescaled by the average and standard deviation of all the elements in the batch that correspond to this convolutional layer.
\\
If we define the average and standard deviation per layer $\alpha$ and given batch $\mathcal{B}$ as 
\begin{equation}\label{eq:mu}
\mu^{\alpha} = \frac{\sum_{ijk}x_{ij}^{(k)}}{|N_b|\times |N_h| \times |N_w|},
\end{equation}
\begin{equation}\label{eq:std}{(\sigma^{\alpha})}^{2} = {\frac{\sum_{ijk}\left(x_{ij}^{(k)}-\mu^{\alpha}\right)^2}{{|N_b|\times |N_h| \times |N_w|}}},
\end{equation}

the generical element   $x_{ij}^{(k)}$  is transformed into
$$ \gamma\cdot\frac{x_{ij}^{(k)} - \mu^{\alpha}}{\sigma^{\alpha}}+\beta, $$
where $\gamma$ and $\beta$ are two learnable parameters.

The average $\mu^{\alpha}$ can be seen as a particular weighted average, and thus as a convolution. Indeed, if we apply a kernel $K$ like \eqref{eq:kernel}, of size $(N_h,N_w)$ with each entry equal to $1/(N_h\cdot N_w)$, to a image with elements $ x_{ij}^{(k)}$, we obtain equation \eqref{eq:mu} modulo $1/N_b$. Similarly, we can construct a new set of data whose elements are tensors of sides $(N_h,N_w)$ and whose entries are $(x_{ij}^{(k)}-\mu^{\alpha})^2 $ and we obtain equation \eqref{eq:std}, again modulo $1/N_b.$

\section{Weighted Mean}\label{sec:weighted}

As mentioned at the end of the previous section, both $\mu^{\alpha}$ and $\sigma^{\alpha}$ are a particular case of weighted averages. Therefore, they can  be computed via a fixed convolutional kernel of the same size of the image $(N_h, N_w)$ and with each entry equal to $1/(N_h, N_w).$
\label{sec:motivation}

\begin{figure}
\begin{center}
\includegraphics[width=0.91\textwidth]{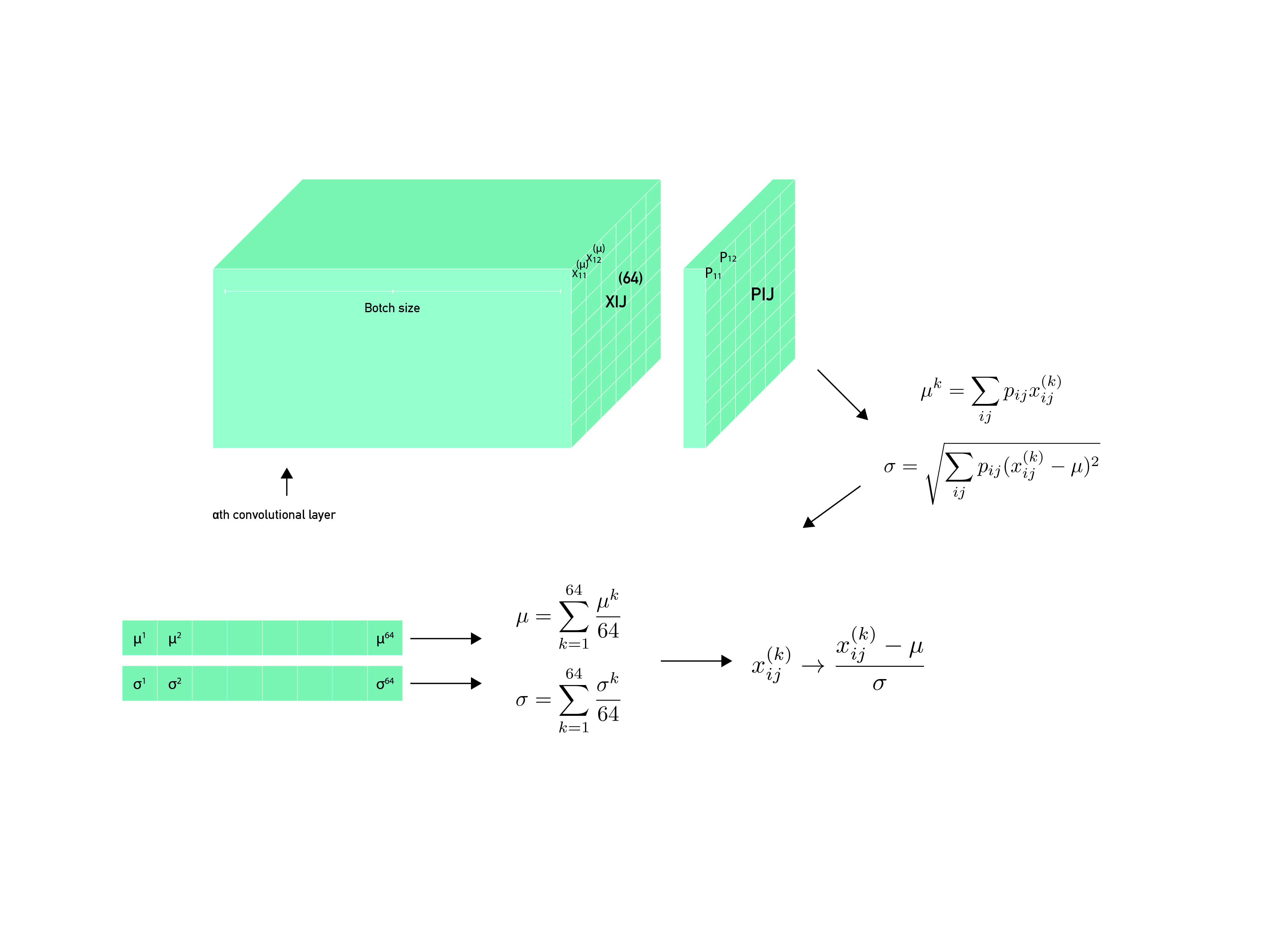}

\caption{A possible generalization of Batch Norm}
\label{fig:CNNN}
\end{center}
\end{figure}

Instead of having weighted sum with predefined coefficients as mentioned at the end of the previous section, we use substitute $K$ with a general kernel, $K_p$ to be learned (see figure \ref{fig:CNNN}).
\\

The weights of the general kernel $K_p$ are chosen such that they satisfy the following constraint:
\begin{align} \label{eq:weighted_mu}
    \sum\limits_{i, j} p_{i, j} = 1, & \quad \text{and} \quad p_{i, j} \ge 0, 
\end{align}
for all $i \in \{1, \dots, N_h \}$ and $j \in \{ 1, \dots, N_w \}$. Thus equations \eqref{eq:mu} and \eqref{eq:std} then become:
\begin{align}\label{eq:weighted_std}
    \mu^{\alpha} = \frac{1}{N_b } \sum\limits_{i, j, k}^{ N_h, N_w, N_b} p_{i, j} x^{(k)}_{i,j },
\end{align}
and similarly, the square of the standard deviation is 
\begin{align}\label{eq:batch_std}
    \left( \sigma^{\alpha}\right)^2 =  \frac{1}{N_b } \sum\limits_{i, j, k}^{N_h, N_w, N_b} p_{i, j}\left( x^{{(k)}}_{i,j} - \mu^\alpha\right)^2.
\end{align}
Notice we have suppressed the index $\alpha$ in the data $x_{ij}^{(k)}$ for the sake of clarity. We again remark that the elements under analysis belong to the given convolutional layer $\alpha.$
\\
We then substitute Batch Normalization with the steps mentioned here using Convolutional Kernel procedure, i.e.
\begin{align}
    \widehat{x_{ij}^{(k)}}  = \frac{x_{ij}^{(k)}  - \mu^\alpha}{\sigma^\alpha},
\end{align}
being this latter a proper generalisation of the former.
\\
\begin{remark}
One justification of this approach is that the arithmetic mean used can lead to false statistics when the data belong to complex spaces. A possible solution is to use importance sampling, where it uses weighted arithmetic mean. The main message is that it can be viewed as a mapping of the data to a different space which is easier to compute the mean and the variance. One thing to point out is the difference between our approach and importance sampling, as the latter uses coefficients determined beforehand while the former learns the coefficients. A more detailed exposition on this argument can be found in Section \ref{sec:importance} of the Appendix. 
\end{remark}
\subsection{Implementation of Weighted Mean}\label{sec:depthwise}
\begin{figure}[htbp]
\centerline{\includegraphics[scale=0.45]{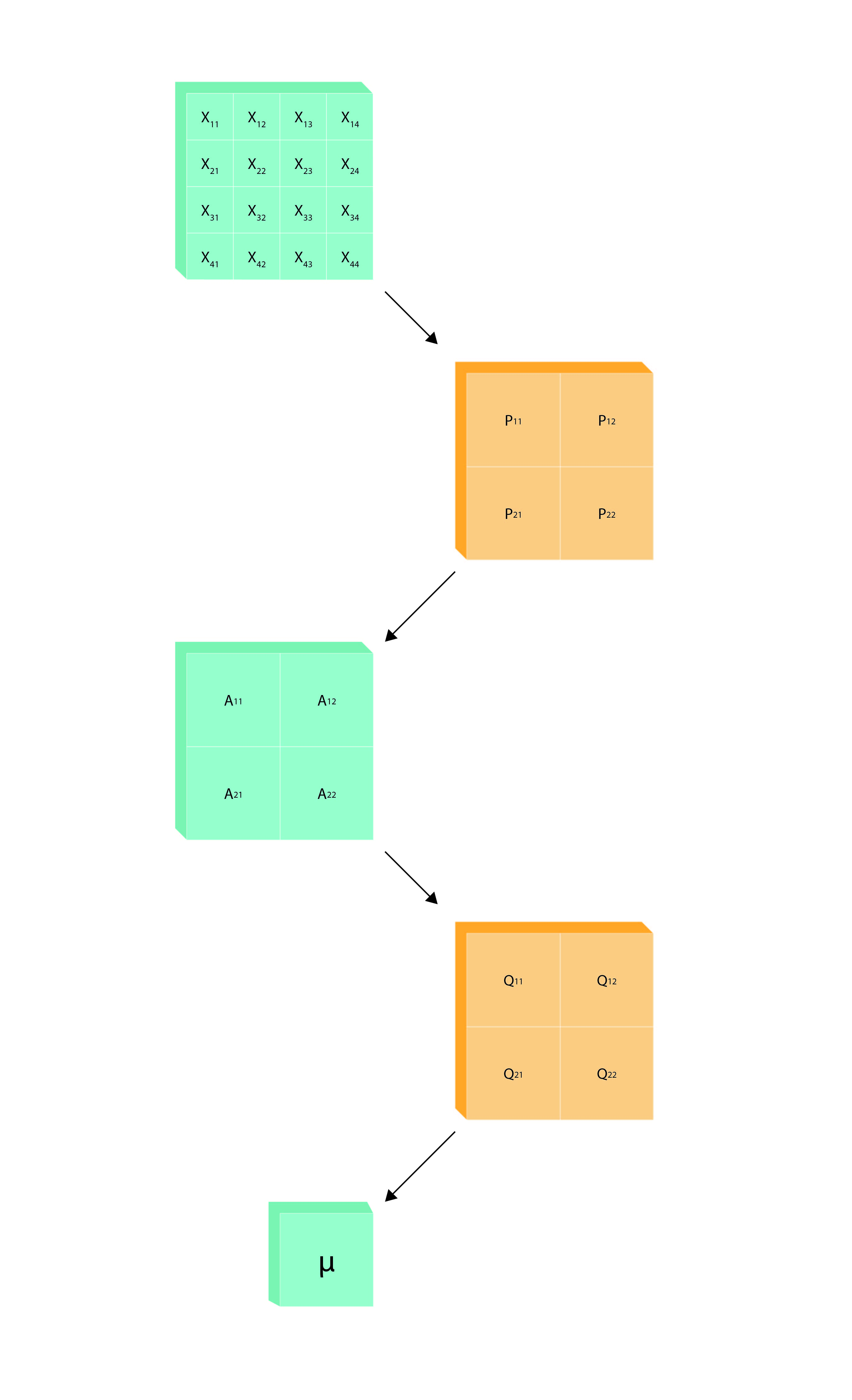} }
\caption{How Conv Norm operates}
\label{fig:ConvNorm}
\end{figure}

The goal is to apply the described convolutional kernel to tensor separately for each feature map. To do this,  we use Depthwise Convolutional Kernels (DWCK henceforth) \cite{DBLP:conf/aaai/GuoLWR19} instead of Convolutional Kernels. 
This latter, indeed, operates along several input channels and is not desirable for our purposes. 
The weights of each DWCK have to be non-negative to verify the second condition of \eqref{eq:weighted_mu}.
Also, using DWCK of the size $height\times width$ can be computationally intensive due to the high numbers of weights we are introducing. To avoid this, we use a stack of several smaller DWCK that emulate a similar result of one kernel of size $height\times width$. In doing this we put attention on the fact we want for each $x_{ij}$ a differnt weight $p_{ij}$. Let's consider the example in the picture \ref{fig:ConvNorm} to see how we can achieve these two latter purposes when we process a $4\times 4$ image:
we initizialy apply a DWCK (in orange in the picture) with kernel size 2x2 and stride 2x2 to the image of shape 4x4 (in green in  the picture).
In this way we get a 2x2 block, that we call A, where each entry of the block contains the following values:
\begin{eqnarray}
\mbox{entry A1:}\qquad p_{11}x_{11}+p_{12}x_{12}+p_{21}x_{21}+p_{22}x_{22},\\
\mbox{entry A2:}\qquad p_{11}x_{13}+p_{12}x_{14}+p_{21}x_{23}+p_{22}x_{24},\\
\mbox{entry A3:}\qquad p_{11}x_{31}+p_{12}x_{32}+p_{21}x_{41}+p_{42}x_{22},\\
\mbox{entry A4:}\qquad p_{11}x_{33}+p_{12}x_{34}+p_{21}x_{43}+p_{22}x_{44}.
\end{eqnarray}
As it can be seen there are some elements that share the same $p-weight.$ For example $x_{11}$ and $x_{13}$.
To avoid this we use another 2x2 DWCK with weights $q$ getting as a final result:
\begin{eqnarray*}
q_{11}\cdot(p_{11}x_{11}+p_{12}x_{12}+p_{21}x_{21}+p_{22}x_{22})+\\
q_{12}\cdot(p_{11}x_{13}+p_{12}x_{14}+p_{21}x_{23}+p_{22}x_{24})+\\
q_{21}\cdot(p_{11}x_{31}+p_{12}x_{32}+p_{21}x_{41}+p_{42}x_{22})+\\
q_{22}\cdot( p_{11}x_{33}+p_{12}x_{34}+p_{21}x_{43}+p_{22}x_{44}).
\end{eqnarray*}
In this way all the $x_{ij}'s$ element have different weights and we have reduced the amount of weights by a factor of 2.
\\
We can compute the standard deviation similarly.
In general we can apply this procedure to all kind of images as far as we use some padding. For further details see the appendix \ref{sec:splitting}.
\subsubsection{Initialising values}
If we have just a DWCK of size $N_h\times N_w$, to be unbiased, we have to set all the entries of the DWCK equal to $1/(N_h \times N_h)$. However, initializing parameters of a neural network to the same value can prevent it from learning. Therefore we uniformly and randomly assign weights to each kernel element in a small neighborhood of $1/(N_h \times N_w)$. Using this methodology, we can both train the net and staying close to our initial hypothesis.
According to the length of the stack of DWCK we use, we might need to change this procedure a bit. In the example in the previous section, we noticed that each $x_{ij}^{(k)} s$ had a different weight, and it was equal to a $p_{ij}$ coming from the first kernel, multiplied by $q_{ab}$ coming from the second kernel.
Therefore, $p_{ij}\cdot q_{ab}$ should be equal to $1/(N_h \times N_w)$. Assuming both $p_{ij}$ and $q_{ab}$ carry the same amount of information we can set each of them equal to $\sqrt{1/(N_h \times N_w)}$.
Indeed if $p_{ij}$ and $q_{ab}$ are similar we have
$$p_{ij}\cdot p_{ij} \sim p_{ij}\cdot q_{ab} = 1/(N_h \times N_w),$$
$$q_{ab}\cdot q_{ab} \sim  p_{ij}\cdot q_{ab} = 1/(N_h \times N_w),$$
which gives the desired result.
If we have longer stacks of DWCK of length $n$ we apply a similar procedure but using $n-th$ root.


\section{Experiments}\label{sec:experiments}

Here we perform a number of experiments to validate the theory and perform quantitative and qualitative analysis of the presented normalization techniques in \ref{sec:motivation}. The experiments are classification and that entails finding a classifier, which can be described mathematically as $g: \mathcal{X} \to \mathcal{C}$ and takes a tensor $x \in \mathcal{X}$ to a label $c \in \mathcal{C}$, where $\mathcal{X}$ denotes the space of inputs of a dataset and for all experiments here $\mathcal{X} \subset \mathcal{B}$ and $\mathcal{C}$ is the space of labels and it is a subset of $\mathbb{R}^{10}$. In most cases the labels are scalar, or text string, but here we use hot-encoding and all experiments have $10$ classes, hence each label is a $10$ dimensional vector.
\\

Since all experiments are classification problems,  the learning paradigm is supervised learning, and it requires the training data set $\{(x_{i_1}, c_{i_1})\}_{i_1}^{N_t}$, the goal is to find a classifier $g$ such taht it minimises the loss functional 
\begin{align}
\mathbb{S} = \sum_{i_1=1}^{N_t} L(g(x_{i_1}), c_{i_1}) 
\end{align}
where $L$ is the categorical cross-entropy defined by
\begin{align*}
L(x_{i_1},c_{i_1})  = - \sum_{i_2}c_{i_1,i_2} \mathrm{log}(x_{i_1, i_2}).
\end{align*}
There are many datasets the can be used for classification problem, but here we limit ourselves the CIFAR-10 dataset \cite{cifar}. For this dataset we use two different architectures, one of which is based on all a classifier known as ALL-CNN-C presented in \cite{DB15a}. The reason for this choice is that the classifier uses solely convolutional neural networks and avoids using any extra layers such as dropout and normalization. And because of that, it presents a good candidate for a classifier where we can study how different normalization methods can improve. The base classifier consists of three two-dimensional CNN with $96$ filters and kernels of the size $3 \times 3$, where the third layer has a $(2,2)$ strides. That followed by four  two-dimensional CNN with $192$ filters and $3 \times 3$ kernel and the third one has a $(2,2)$ strides. Then followed by a two dimensional CNN with $192$ filters with $1 \times 1$ kernel. The penultimate layer is a two dimensional CNN with $10$ filters and $1 \times 1$ kernel and then fed into an average pooling layer, which outputs a $10$ dimensional vector. Mathematically, the classifier is written as 
\begin{equation}\label{eq:classifier}
\begin{aligned}
x^l_{i_1,i_2} &= (\phi \circ f^l_{N}) \left( w^l_{i_2} \ast x^{l-1}_{i_1, i_2} + b^l_{i_2} \right),  
\end{aligned}
\end{equation}
for $l =1, \dots, N_h$ to denote the layer and $w^l_{i_2}$ and $b^l_{i_2}$  are the kernel and bias, respectively, of the $i_2^{th}$ feature map of the $l^{th}$ layer. The index $i_1 = 0, \dots, N_b$ denotes the $i_1^{th}$ element of $\mathcal{X}$.
The $f_{N}$ denotes the normalizing method. Starting with Batch Normalization, reviewed in section \ref{sec:background}, the normalization method $f_N$ is here denoted by $BN$. The method does not use neural networks per se, but it learns two parameters, $\gamma$ and $\beta$, for each feature. This renders the method with a low number of parameters. 

In addition to Batch Normalization, we apply the a weighted normalization method, described in section \ref{sec:weighted}, where the weighted sum is implemented using depth-wise convolutional neural networks. here we can set $f_{N} = DWCK(a,b)$, where $a$ is the kernel size and $b$ is the stride. The normalization for the first two layers of ALL-CNN-C network is implemented using four successive layers of depth-wise CNN. For the third, fourth and fifth layers of ALL-CNN-C, the normalization consists of three layers of depth-wise CNN. For the sixth layer, two layers of depth-wise CNN are utilised. The dimensions and strides of the DWCK are a mix of (4,4) and (2,2) as shown in the table below.

The following table summarises the the classifier used for the numerical experiments presented in this section. One thing to note, in order to include as much as information in the table, we use the notation $(c, a, b)$ to indicate the $c$ number of channels in the CNN with kernel of size $a$ and stride $b$. Whenever the third number is not included, it means the stride is set to $1$. 
\begin{center}
\begin{tabular}{ |c|c|c|c| } 
\hline
\multicolumn{3}{|c|}{Classifier} \\
\hline
ALL-CNN-C & ALL-CNN with BN & ALL-CNN with DWCK \\
\hline
$(96, 3\times 3)$ with $ReLU$ & $(96, 3\times 3)$  with $ReLU$& $(96, 3\times 3)$  with $ReLU$ \\ 
& $f^1 = BN$ & $f^1 = 1\ DWCK(4,4)$ \\ 
& &$ + 3 DWCK(2,2)$ \\ 
\hline
$(96, 3\times 3)$  with $ReLU$ & $(96, 3\times 3)$ with $ReLU$ & $(96, 3\times 3)$ with $ReLU$ \\ 
& $f^2 = BN$ & $f^2 =1\ DWCK(4,4)$ \\ 
& &$ + 3 DWCK(2,2)$ \\ 
\hline
$(96, 3\times 3,2)$ with $ReLU$ & $(96, 3\times 3,2)$ with $ReLU$ & $(96, 3\times 3,2)$ with $ReLU$ \\ 
& $f^3 = BN$ & $f^3 = 1\ DWCK(4,4)$ \\ 
& &$ + 2 DWCK(2,2)$ \\ 
\hline
$(192, 3\times 3)$ with $ReLU$ & $(192, 3\times 3)$ with $ReLU$ & $(192, 3\times 3)$ with $ReLU$ \\ 
& $f^4 = BN$ & $f^4 = DWCK(4,4)$ \\ 
& &$ + 2 DWCK(2,2)$ \\ 
\hline
$(192, 3\times 3)$ with $ReLU$ & $(192, 3\times 3)$ with $ReLU$ & $(192, 3\times 3)$ with $ReLU$ \\ 
& $f^5 = BN$ & $f^5 =1\  DWCK(4,4)$ \\ 
& & $+ 2 DWCK(2,2)$ \\ 
\hline
$(192, 3\times 3,2)$ with $ReLU$ & $(192, 3\times 3,2)$ with $ReLU$ & $(192, 3\times 3,2)$ with $ReLU$ \\ 
& $f^6 = BN$ & $f^6 =1\ DWCK(2,2)$ \\ 
& & $+ 1 DWCK(2,2)$ \\ 
\hline
$(192, 3\times 3)$ with $ReLU$ & $(192, 3\times 3)$ with $ReLU$ & $(192, 3\times 3)$ with $ReLU$ \\ 
& $f^7 = BN$ & $f^7 = 1\ DWCK(2,2)$ \\ 
& &$ + 1 DWCK(2,2)$ \\ 
\hline
$(192, 1\times 1)$ with $ReLU$ & $(192, 1\times 1)$ with $ReLU$ & $(192, 1\times 1)$ with $ReLU$ \\ 
& $f^7 = BN$ & $f^7 = DWCK(2,2)$ \\ 

\hline
\multicolumn{3}{|c|} {$(10, 1\times 1)$ } \\
\hline
\multicolumn{3}{|c|} {$\mathrm{Average Pooling}$ with $\mathrm{SoftMax}$ }\\
\hline
\end{tabular}
\end{center}

The results for ALL-CNN with BN and ALL-CNN with DWCK were comparable for batch sizes of $256, 128,$ and $64$.
This for the learning rates we have tried: $0.1, 0.05, 0.01, 0.005, 0.001.$ However, for most of the learning rates for batch sizes of $32$ and $16$ ALL-CNN with DWCK resulted superior and more stable (see pics below) and in appendix \ref{sec:resultspics}. It is also worth mentioning that we used a simple standard deviation since a weighted one was computationally unstable.
\begin{figure}[ht]
\centerline{\includegraphics[scale=0.5]{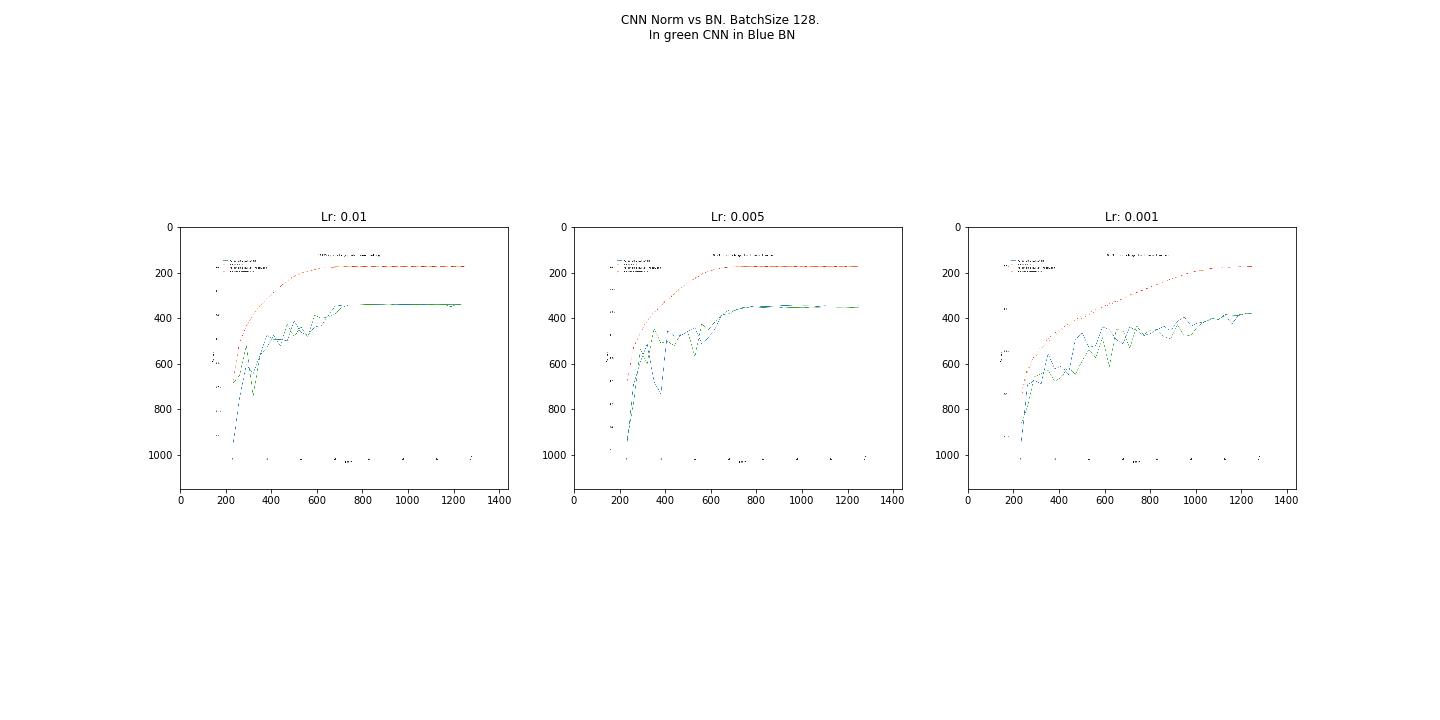} }
\end{figure}
\begin{figure}[ht]
\centerline{\includegraphics[scale=0.35]{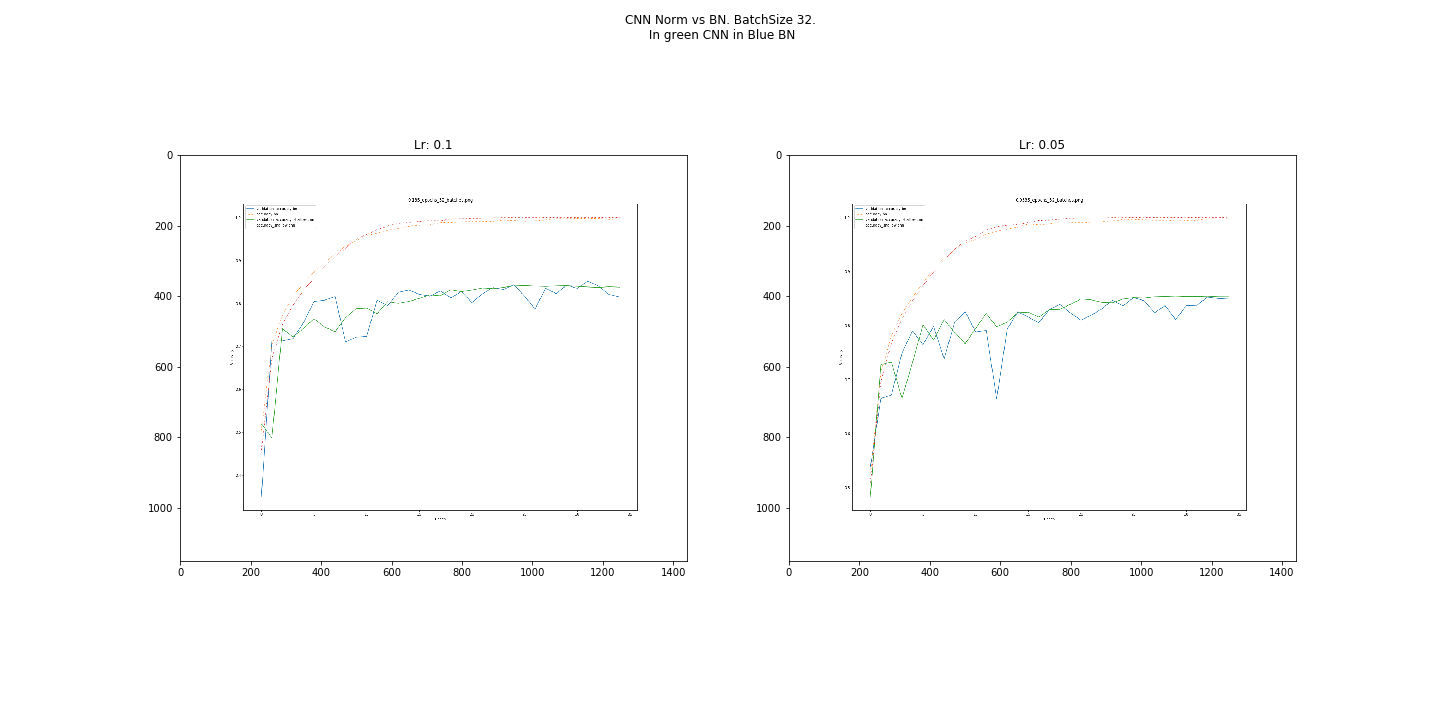} }
\end{figure}
\begin{figure}[ht]
\centerline{\includegraphics[scale=0.50]{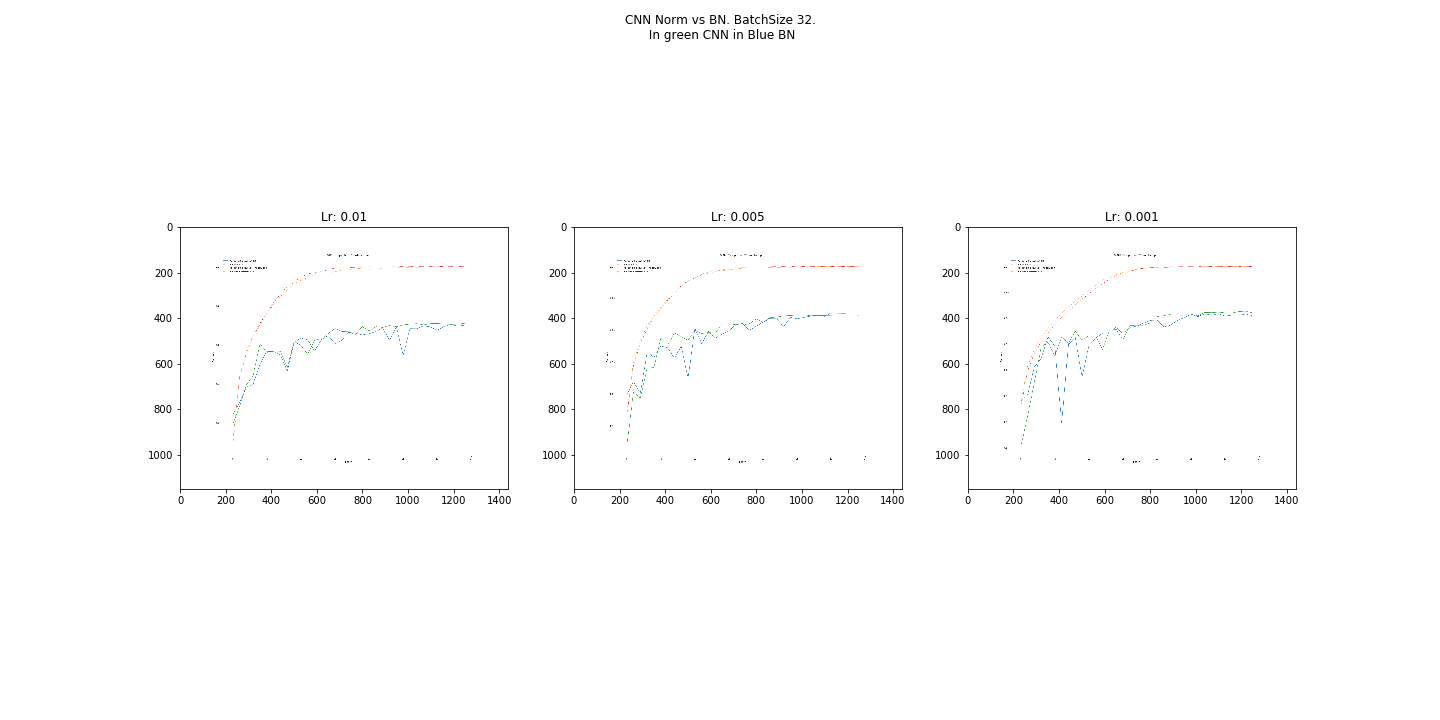} }
\end{figure}
\\

\section{Normalization Using Learned Statistics}\label{sec:inference_nn}
In the previous sections, there are two issues that arise: the first, concerns the number of extra parameters added to the classifier thanks to DWCK is large in comparison to Batch Normalization and that could be a potential issue if bigger DWCK and smaller hardware memory are used. The second issue concerns the use of an arithmetic sum to approximate the mean and standard deviation, it does not take into consideration the space in which the data lives and thus can lead to statistics that do not reflect the true ones.  Furthermore, for it to give an accurate approximation, it requires a training dataset with a very large number of elements and in some application there are challenges with the availability of such training datasets  \cite{mackay2003information}.  
\\

One possible way to address the issues mentioned, is to consider a normalization method with the mean and standard deviation represented using neural networks. This in turn gives freedom of choosing a suitable architecture with a significantly fewer number of parameters. This fact is demonstrated in section \ref{sec:num_nn_experiments}, where such normalization method is shown its ability to accelerate training using a fraction of the parameters that Batch Normalization would require. That was achieved using two  networks each composed of global average pooling layer, so that the result is a vector representing the mean of each channel, and a series of one-dimensional convolutional neural layers followed by another average pooling layer.
\\

Regarding the issue with approximating the mean and the standard deviation, one way is to use a method that is inspired by Bayesian inference. The reason for that is because Bayesian methods have the ability to approximate the parameters of distribution without the need of a large training set, and obtain statistics from data with non-Gaussian distribution \cite{jordan1999introduction,beal2003variational,krishnan2017structured}. One obstacle to applying many Bayesian inference techniques is that they suffer from intractability issues, meaning some quantities cannot be computed. However, neural networks can be used to instead learn the intractable quantities \cite{kingma2013auto}. The same two networks mentioned above are used to approximate the mean and the standard deviation.

\subsection{Numerical Experiments}\label{sec:num_nn_experiments}
Similar to section \ref{sec:experiments}, we add normalizing layers that use CNN neural networks to learn the statistics to ALL-CNN-C classifier. For the first layer of ALL-CNN-C, the networks for both the mean and standard deviation are implemented by first applying a global averaging pool layer, then followed by three one-dimensional CNN layers with $4$, $4$, and $1$ channels respectively and kernel sizes of $4$, $3$ and $3$ respectively. Then followed by an average pooling layer. The reason for having a global pooling layer is to reduce the dimensionality of the input, i.e. $x \in I $ is mapped to a vector in $\mathbb{R}^{N_f}$, where $N_f$ is the number of features maps of $x$. The remaining layers of ALL-CNN-C all have their statistics parametrised by a sequence of global average pooling followed by four one-dimensional CNN layers with $4$, $4$, $2$ and $1$ channels and each having kernels with sizes $4$, $3$, $2$ and $3$, respectively. Similar to the normalization used for the first layer in ALL-CNN-C, an average pooling layer is used as a final layer. 
\\

\begin{figure}[ht]
\centering
\includegraphics[width=0.85\textwidth]{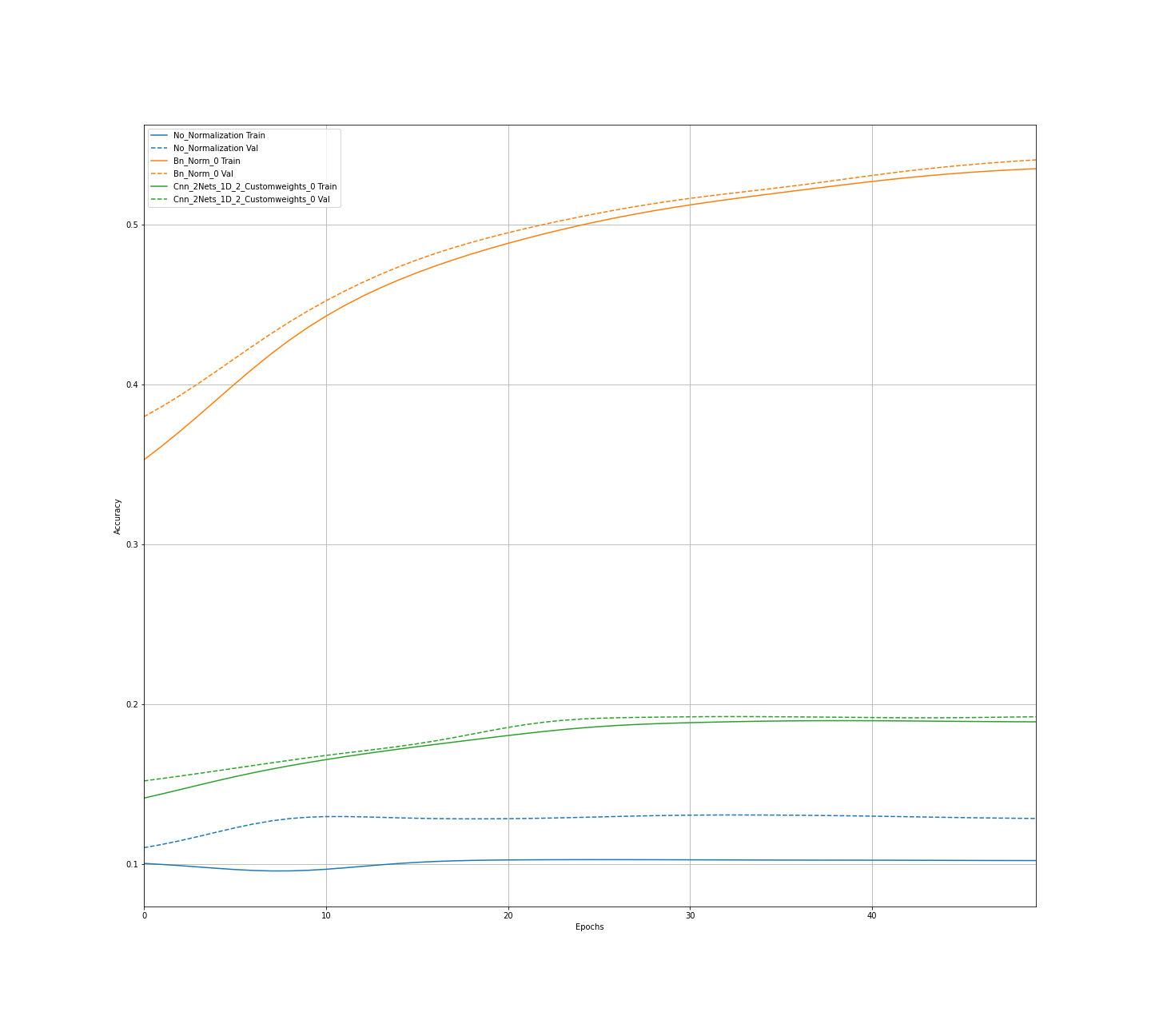}
\caption{(This one I have to redo it) In this figure, the result of running CIFAR $10$ classification for $50$ epochs and with a learning rate set to $10^{-4}$ and the classifiers used are Batch Normalization and one-dimensional CNN's. The dashed lines are the accuracy of the validation set and the solid lines are the training set, while the blue coloured lines are the results plain ALL-CNN-C classifier, the blue lines for Batch Normalized classifier and the green lines are for the learned classifier. }
\label{fig:accuracy_0001}
\end{figure}

When used with CIFAR10 classification problem, described in section \ref{sec:experiments}, in cases with a learning rate in the range of $[10^{-5},10^{-3}]$, this normaliziation approach does not match the validation accuracy of Batch Normalization, yet making gains with respect to the plain classifier. An example of such behaviour can be seen in Figure \ref{fig:accuracy_0001}.
\\

\begin{figure}[ht]
\centering
\includegraphics[width=0.85\textwidth]{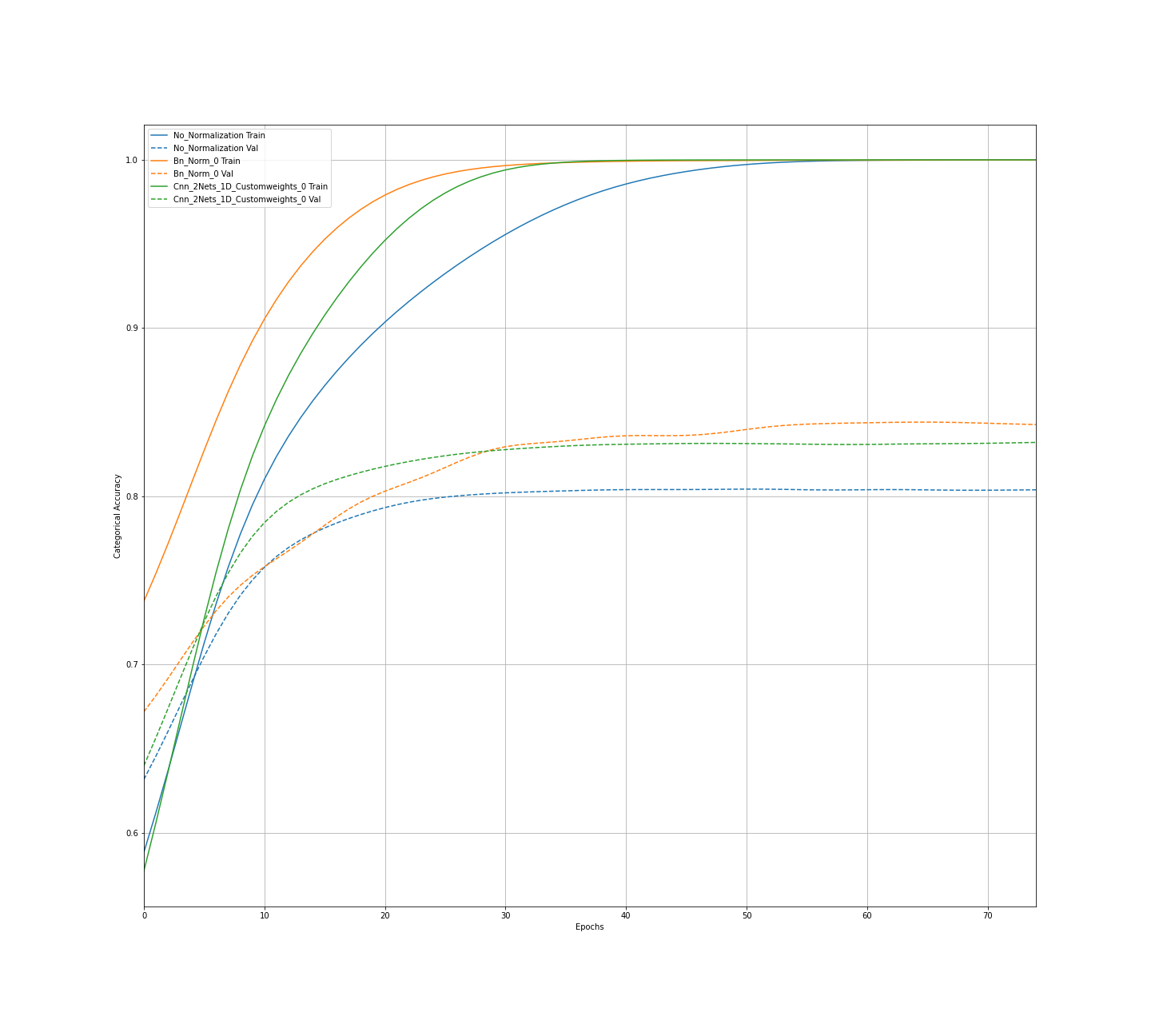}
\caption{In this figure, the result of running CIFAR $10$ classification for $75$ epochs and with a learning rate set to $0.025$ and the classifiers used are Batch Normalization and one-dimensional CNN's. The dashed lines are the accuracy of the validation set and the solid lines are the training set, while the blue coloured lines are the results plain ALL-CNN-C classifier, the blue lines for Batch Normalized classifier and the green lines are for the learned classifier.}
\label{fig:accuracy_0025}
\end{figure}

\begin{figure}[ht]
\centering
\includegraphics[width=0.85\textwidth]{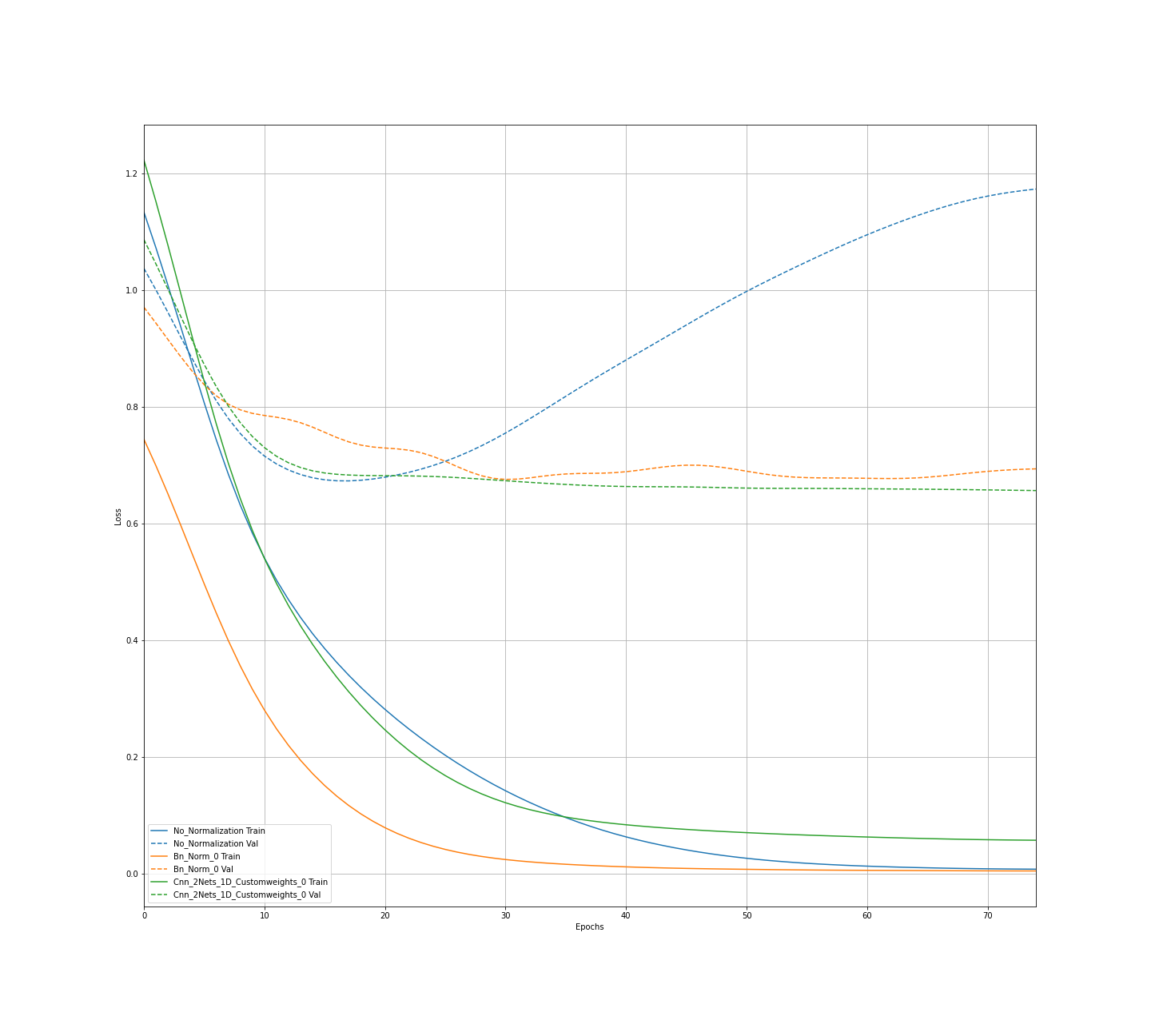}
\caption{In this figure, the result of running CIFAR $10$ classification for $75$ epochs and with a learning rate set to $0.025$ and the classifiers used are Batch Normalization and one-dimensional CNN's. The dashed lines are the loss of the validation set and the solid lines are the training set, while the blue coloured lines are the results plain ALL-CNN-C classifier, the blue lines for Batch Normalized classifier and the green lines are for the learned classifier.}
\label{fig:loss_0025}
\end{figure}

However, when the learning rate is in the range $[0.05, 0.01]$, we notice that the method initially out performs Batch Normalization when it comes to validation accuracy and validation loss. This is shown to be the case in Figure \ref{fig:accuracy_0025} where the accuracy of the CNN normalization is noticibly higher than than Batch Normalization for the first $25$ epochs. However, for the same classifiers, the loss of the CNN normalization is lower and less oscillating than Batch Normalization throughtout the duration of the training as shown in Figure \ref{fig:loss_0025}.

\section{Conclusion}
We have presented a normalizing method that helps increasing the convergence rate of deep convolutional neural network training. At the core of the method a weighted arithmetic sum that is used to compute the mean and the standard deviation used for centring and scaling the output of the hidden layers of the  deep convolutional neural network. Usually  weighted arithmetic sum has fixed weights, but here the weights are not fixed and instead they are learned from the data. This is realised using depth-wise convolutional neural networks, whose kernel weights fulfil the role of the weights of the arithmetic sum.  The experimental results on CIFAR10 dataset show that his method can indeed match the performance of Batch Normalization, which here is used as a benchmark. 
\section{Future Work}
Thus far we explored normalization techniques which can be summarised as mapping the images of a dataset to another set whose images all have a mean zero and standard deviation equal to one. However this does not result in a decorrelated data and to address this, in \cite{8578187}, a Zero Phase Component Analysis (ZCA) was introduced to Batch Normalization process. However, ZCA requires a number of linear algebra operations, including inverting a matrix, which can be a potential bottleneck when it comes to higher dimensional data. One solution was introduced in \cite{huang2019iterative}, where an iterative scheme was used to approximate the inverse of the covariance matrix. One avenue of research is to investigate the whether the addition of decorrelation step to DWCK described in section \ref{sec:weighted} can yield improved results. 
\\

Another research question one can ask is whether or not the addition of normalizing flows of \cite{rezende2015variational} can help improve the performance of our method. A Normalizing flow, 
in short, is a series of simple invertible transformations applied iteratively to an image and it learns to map an image into a two dimensional simple distribution, such as the Normal distribution. This is helpful as the mean and standard deviation of the normal distribution are known, so we can use the inverse mapping to obtain the mean and standard deviation of an image. To answer why this is a subject of interest is due to the ability of normalizing flows to map between simple distribution and complex non-Gaussian ones. 
\\

Finally, we aim to see if it is possible to resolve the computational problems of introducing weighted standard deviation in our first methodology of section \ref{sec:weighted}. This could indeed enhance further our results.

\bibliography{main.bib}


\bibliographystyle{plain}
\newpage
\appendix
\section{Importance Sampling View of DWCK}\label{sec:importance}
Treating the hidden layers' activations as independent and identically distributed random variables $X^{\mu}: (\Omega, \mathcal{F}, \mathbb{P}) \to \mathbb{R}^{N^l_h}_{+}$, where ${N^l_h}$ is the number of nodes in the $l^{th}$ layer and $ (\Omega, \mathcal{F}, \mathbb{P})$ is the probability space adapted to filtration $\mathcal{F}$ and with probability space $\mathbb{P}$. 

To justify our approach, we look towards statistics, as the method used to compute the statistics is through Monte Carlo method and according to the Law of Large Numbers, i.e. $N_b \to \infty$, we have
\begin{align}\label{eq:monte_carlo_mean}
\widehat{\mu} = \frac{1}{N_b} \sum\limits_{i=0}^{N_b} X_i \to \mathbb{E}_{\mu}[x] = \int x \, d\mu = \int x \rho(x) dx, 
\end{align}
where $\rho(x)$ is the probability density function and $\mu$ is the probability measure on the probability space $\mathbb{P}$. In a way, Monte Carlo method can be viewed as Riemann summation of the continuous integral on \eqref{eq:monte_carlo_mean} with $\rho(x)$ set to unity for all $x$. And similar to other quadrature methods, Monte Carlo method fails to approximate the mean of random variables whose underlying probability distribution is difficult to sample from and it leads to having a numerical method with a very large variance. In order to decrease the variance and obtain an approximation close to the true mean, we sample from a different probability space $\mathbb{Q}$ with probability measure $d\nu$ and not $\mathbb{P}$.
\\

Consider the Brownian motions $B_t$ and $Y_t = B_t + \theta \, t$, where $\theta$ is the drift of the stochastic process $Y_t$. According to Girsanov's theorem \cite{doi:10.1137/1105027}, the processes $B_t$ and $Y_t$ are equivalent when the probability space $\mathbb{P}$ is changed to $\mathbb{Q}$. In other words, Girsanov's theorem tells us how random variables on probability space $(\Omega, \mathcal{F}, \mathbb{P})$ are defined when the probability space is changed to $\mathbb{Q}$. One manifestation of Girsanov's theorem is the method of importance sampling, which is used in the numerical calculation of expected values. This takes us back to \eqref{eq:monte_carlo_mean}, we insert $\frac{\xi(x)}{\xi(x)}$ in the integrand, where $\xi(x)$ is a probability density function of the probability distribution of $\mathbb{Q}$, and we obtain 
\begin{align}
    \mathbb{E}_{\nu}[x] = \int x \rho(x) \frac{\xi(x)}{\xi(x)} dx = \int x  \frac{\rho(x)}{\xi(x)} \xi(x) dx = \mathbb{E}_{\mu} \left[x\frac{\rho(x)}{\xi(x)}\right]
\end{align}
or in terms of measures
\begin{align*}
    \mathbb{E}_{\nu}[x] = \int x \frac{d \mu}{d \nu} d\nu.
\end{align*}
The term $\frac{d\mu}{d\nu}$ is known as the \textit{Radon-Nikodym derivative} \cite{mackay2003information}. 
With that, the approximation of the mean is 
\begin{align}\label{eq:imp_sampling}
    \widehat{\mu} = \frac{1}{N_b} \sum\limits_{i= 0}^{N_b} x_i p_i \approx \mathbb{E}_{\nu} [ x],
\end{align}
where the coefficients $p_i = \frac{\rho(x_i)}{\xi(x_i)}$ and $x \simeq \mathbb{Q}$ and this is known as \textit{importance sampling} \cite{mackay2003information}. For the case of the stochastic processes $B_t$ and $Y_t$, the Radon-Nikodym derivative is computed as $\mathrm{exp}\left( - x \mu + \frac{\mu^2}{2} \right)$. Notice that equation \eqref{eq:imp_sampling} is the same as equation \eqref{eq:weighted_std}.

\section{Details of section \ref{sec:depthwise}}\label{sec:splitting}
Before explaining the procedure in general, let's examine another example. Let's assume we have an image $I$ $32\times32$. We can factor $32$ in $4\times2\times2\times2$. We can stack DWCK of sizes $(4,4)$, $(2,2)$, $(2,2)$, $(2,2)$, with strides respectively $(4,4)$, $(2,2)$, $(2,2)$, $(2,2)$, If we let pass $I$ through the stack of DWCK we get a single number which can be interpret, according to the cases, as the sandard deviation or the mean.
In general if we have an immage $I$ of sides $a\times b$ we can factor $a$ and $b$ as $a = a_1\times\dots\times a_n$ and $a_1\times\dots\times a_n$ (if $a$ and $b$ are primes we can pad the image). We can then construct DWCK stack of kernels $(a_1,b_1),\dots(a_n,b_n)$ and strides $(a_1,b_1),\dots(a_n,b_n)$, getting again a single value which can be itnerpreted as the mean or standard deviation.
\section{Results Weighted Mean First Architecture CIFAR 10}\label{sec:resultspics}

\begin{figure}[htbp]
\centering
\includegraphics[width=1\textwidth]{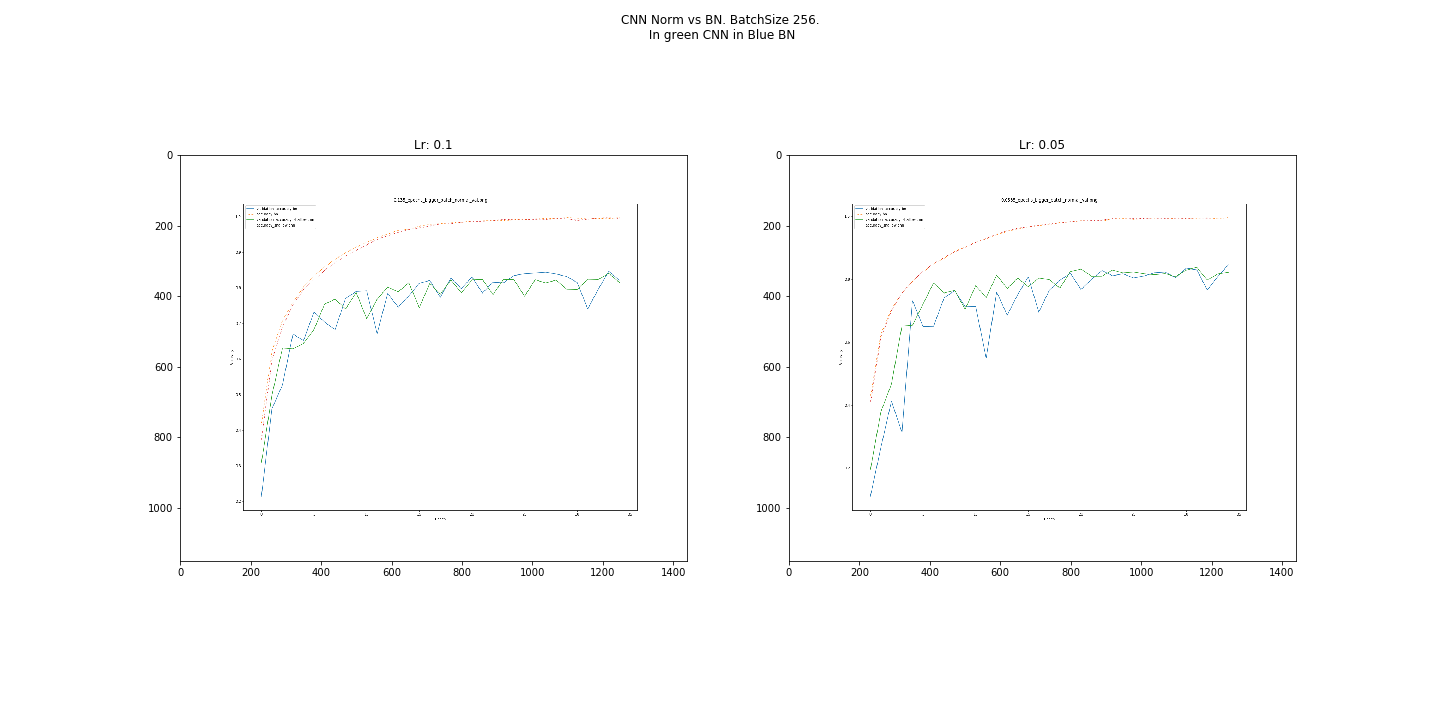}
\end{figure}

\begin{figure}[htbp]
\centering
\includegraphics[width=1\textwidth]{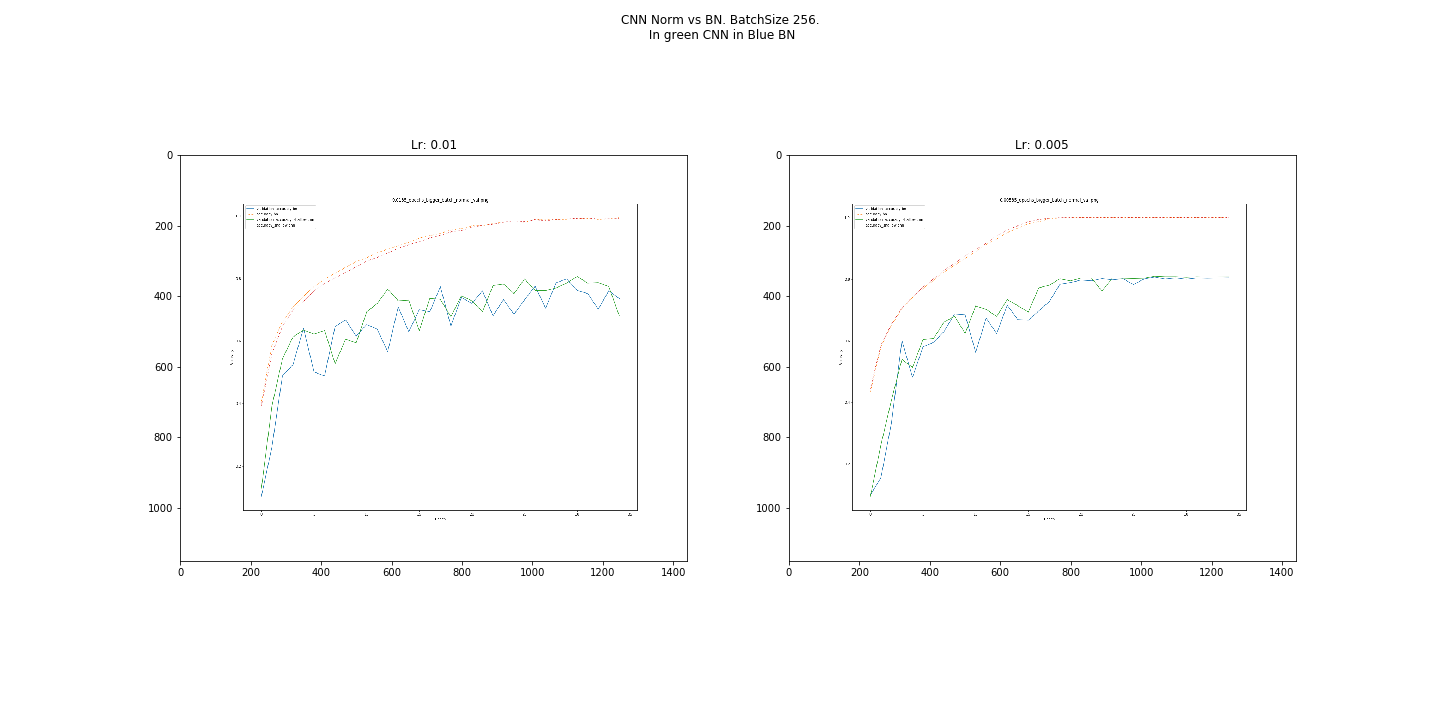}
\end{figure}

\begin{figure}[htbp]
\centering
\includegraphics[width=1\textwidth]{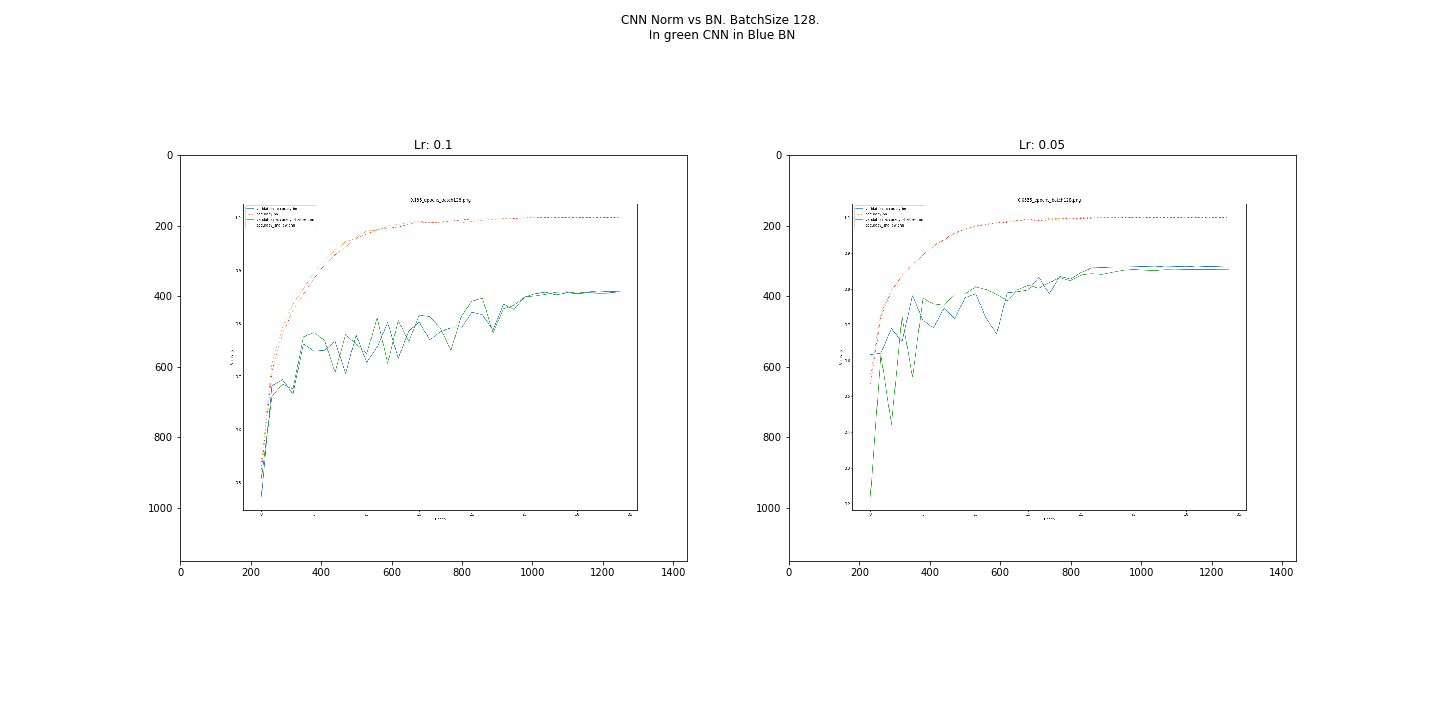}
\end{figure}

\begin{figure}[htbp]
\centering
\includegraphics[width=1.\textwidth]{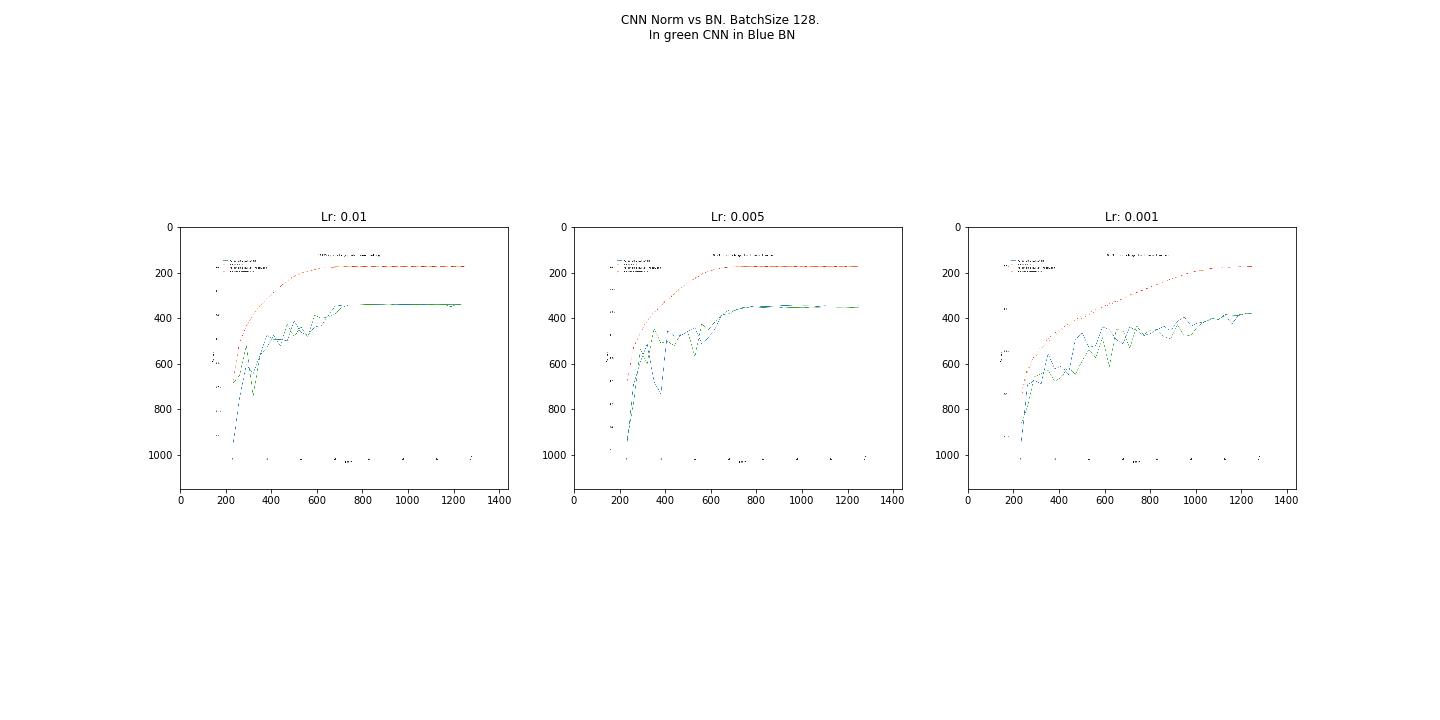}
\end{figure}

\begin{figure}[htbp]
\centering
\includegraphics[width=1\textwidth]{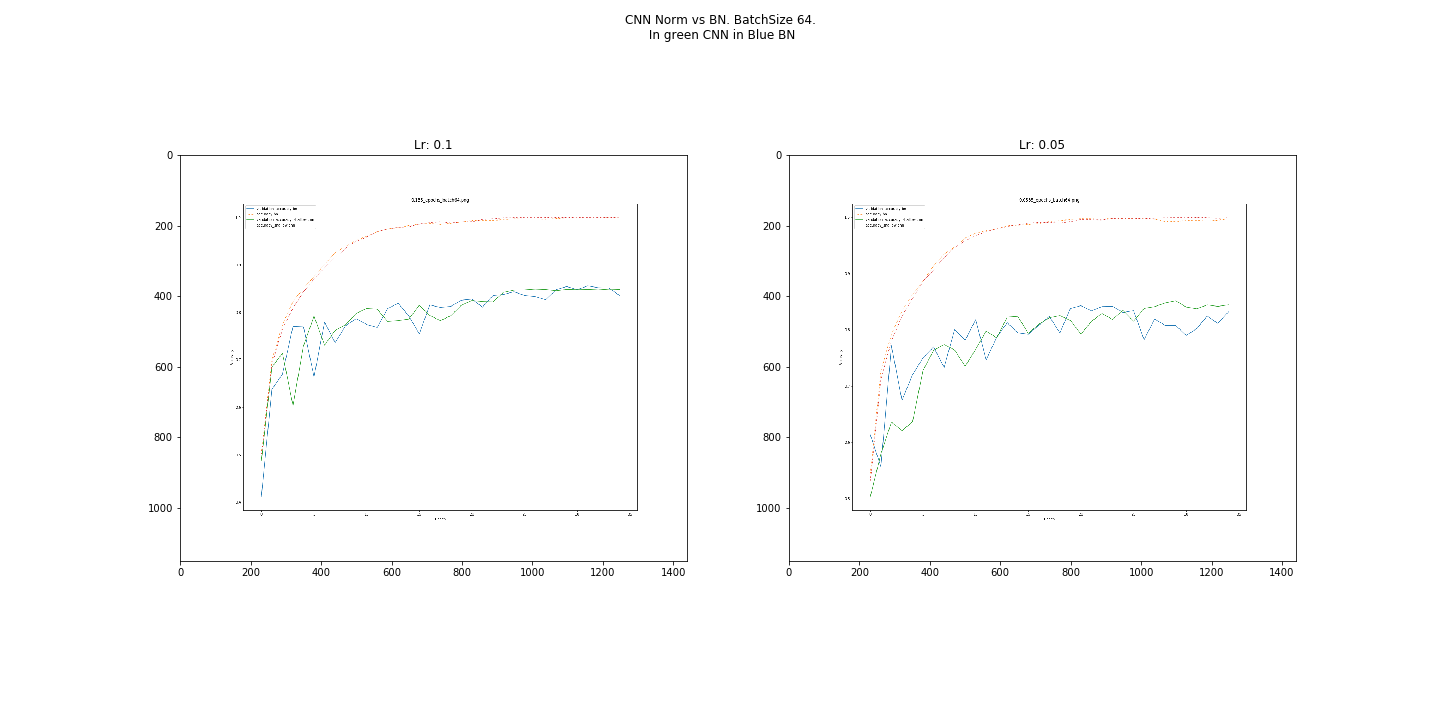}
\end{figure}

\begin{figure}[htbp]
\centering
\includegraphics[width=1.\textwidth]{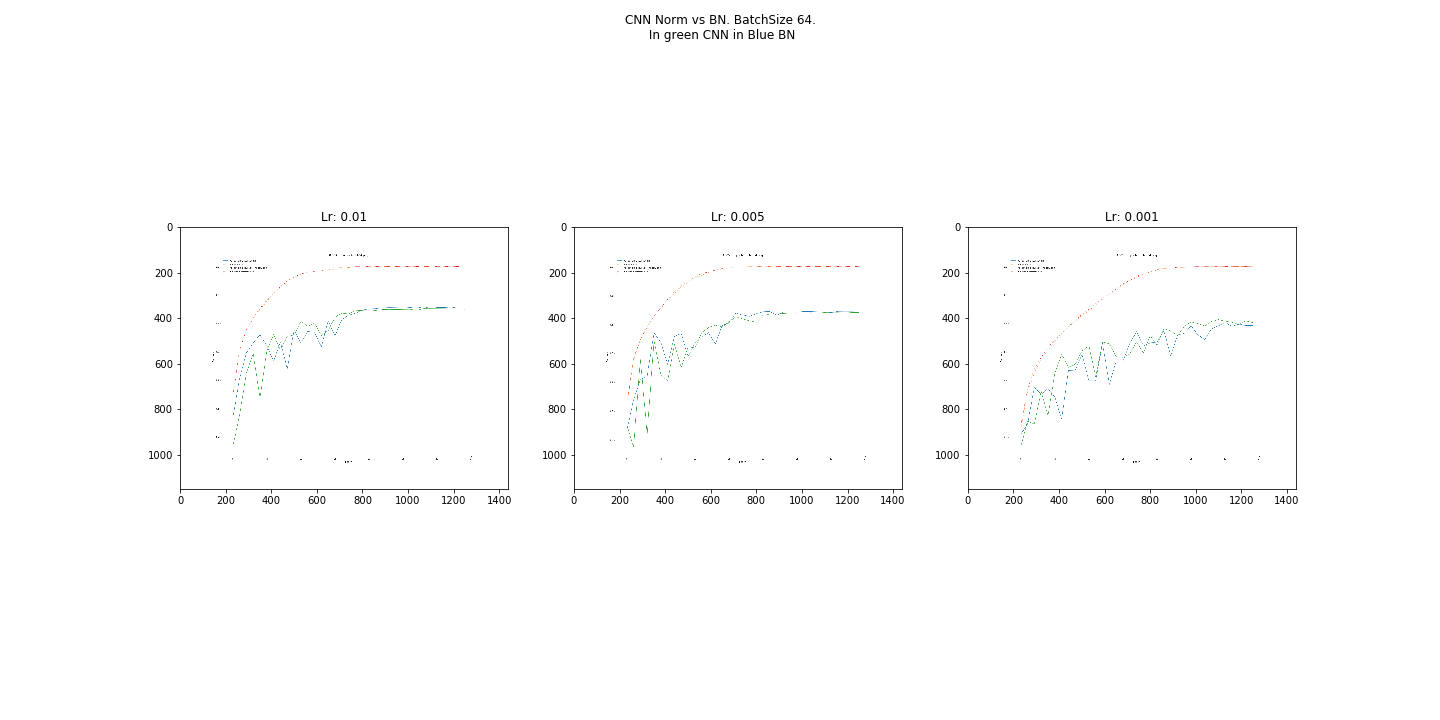}
\end{figure}

\begin{figure}
\centering
\includegraphics[width=1\textwidth]{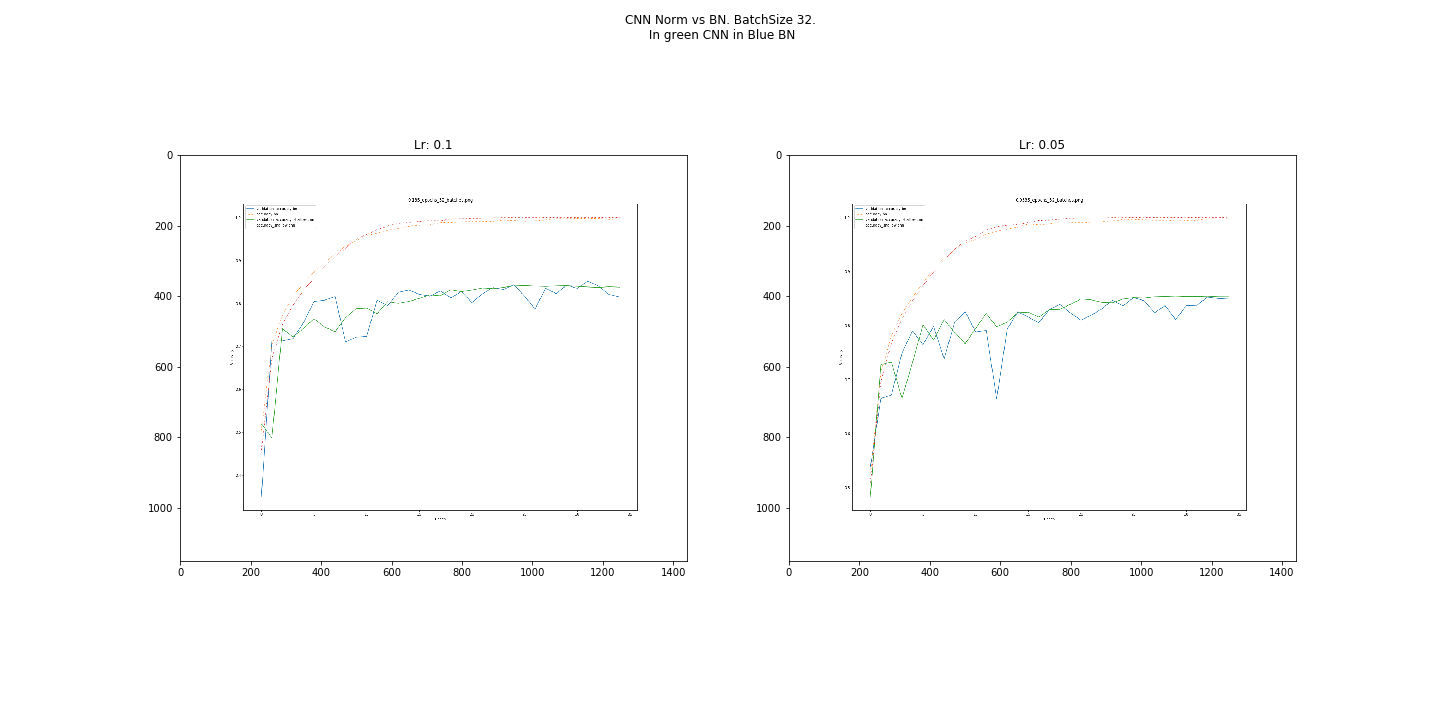}
\end{figure}

\begin{figure}[htbp]
\centering
\includegraphics[width=1.\textwidth]{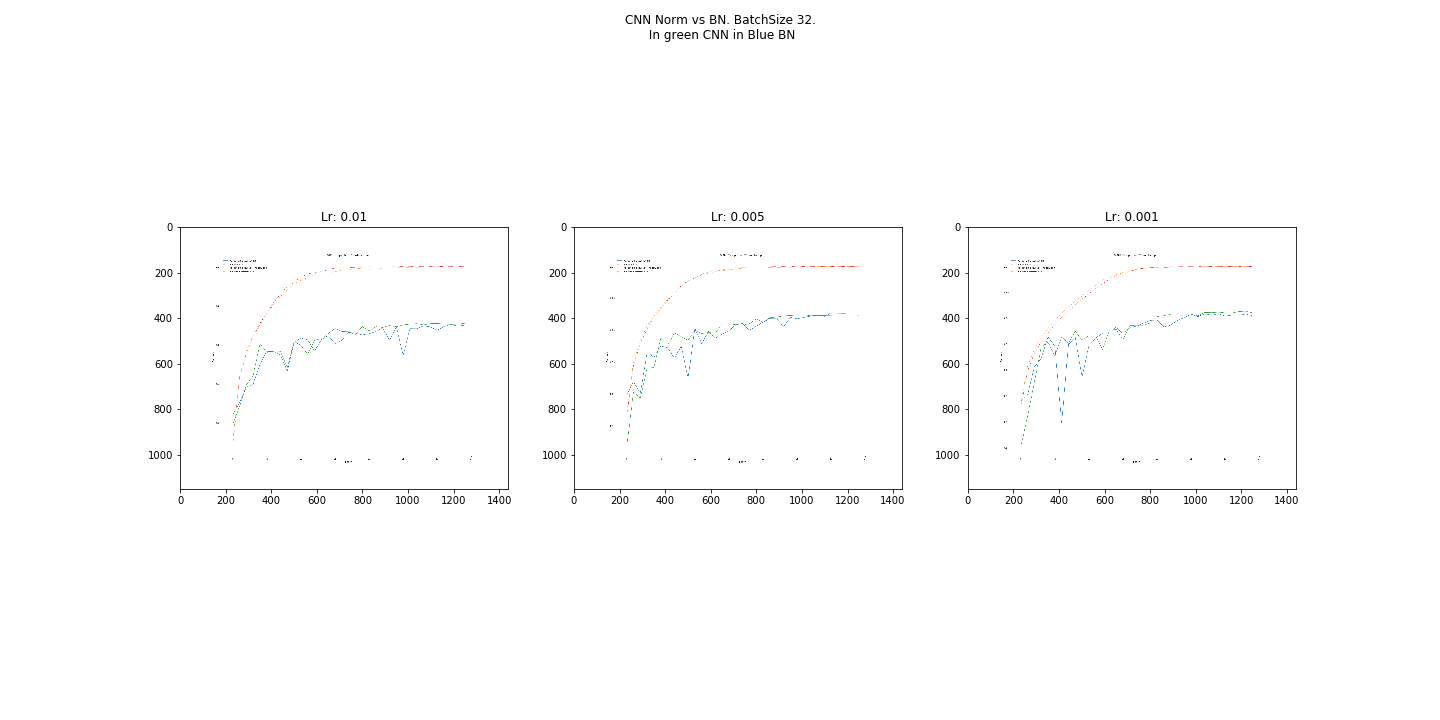}
\end{figure}

\begin{figure}
\centering
\includegraphics[width=1\textwidth]{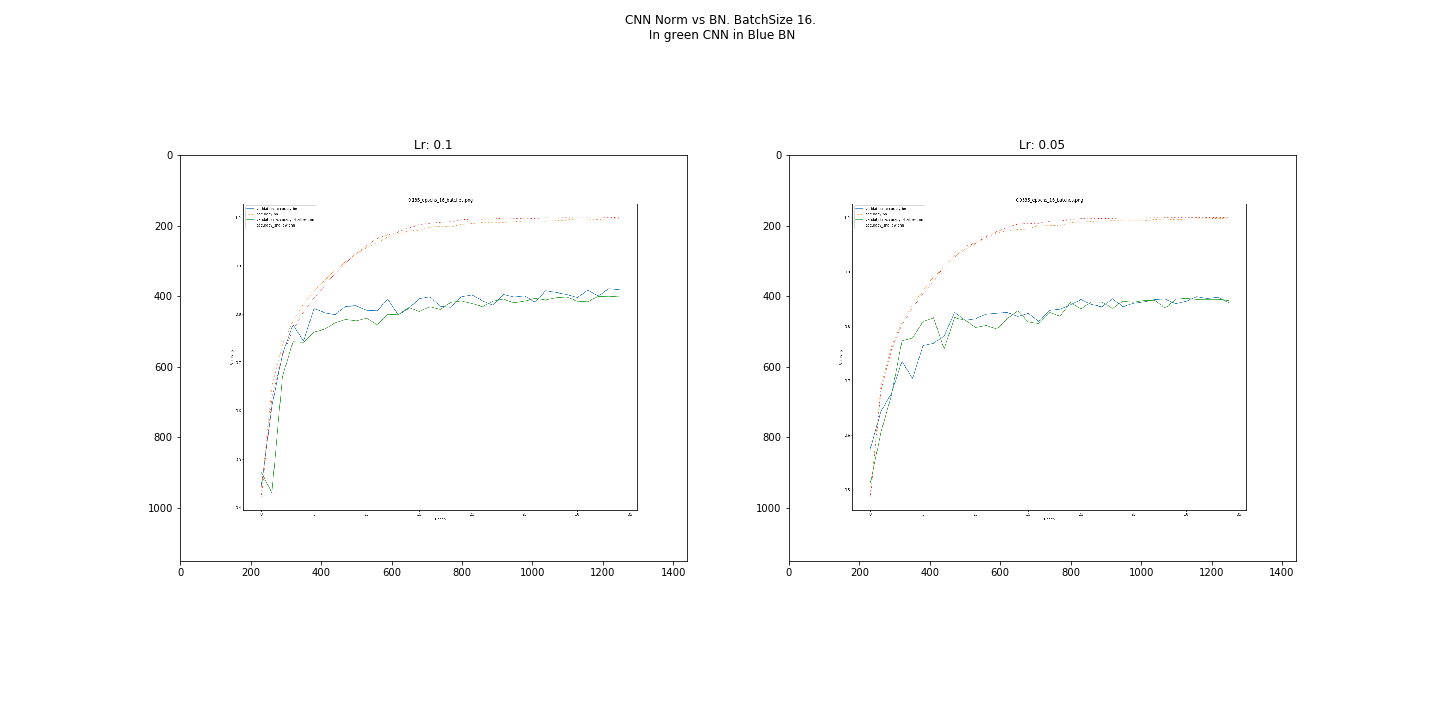}
\end{figure}
\begin{figure}[htbp]
\centering
\includegraphics[width=1.\textwidth]{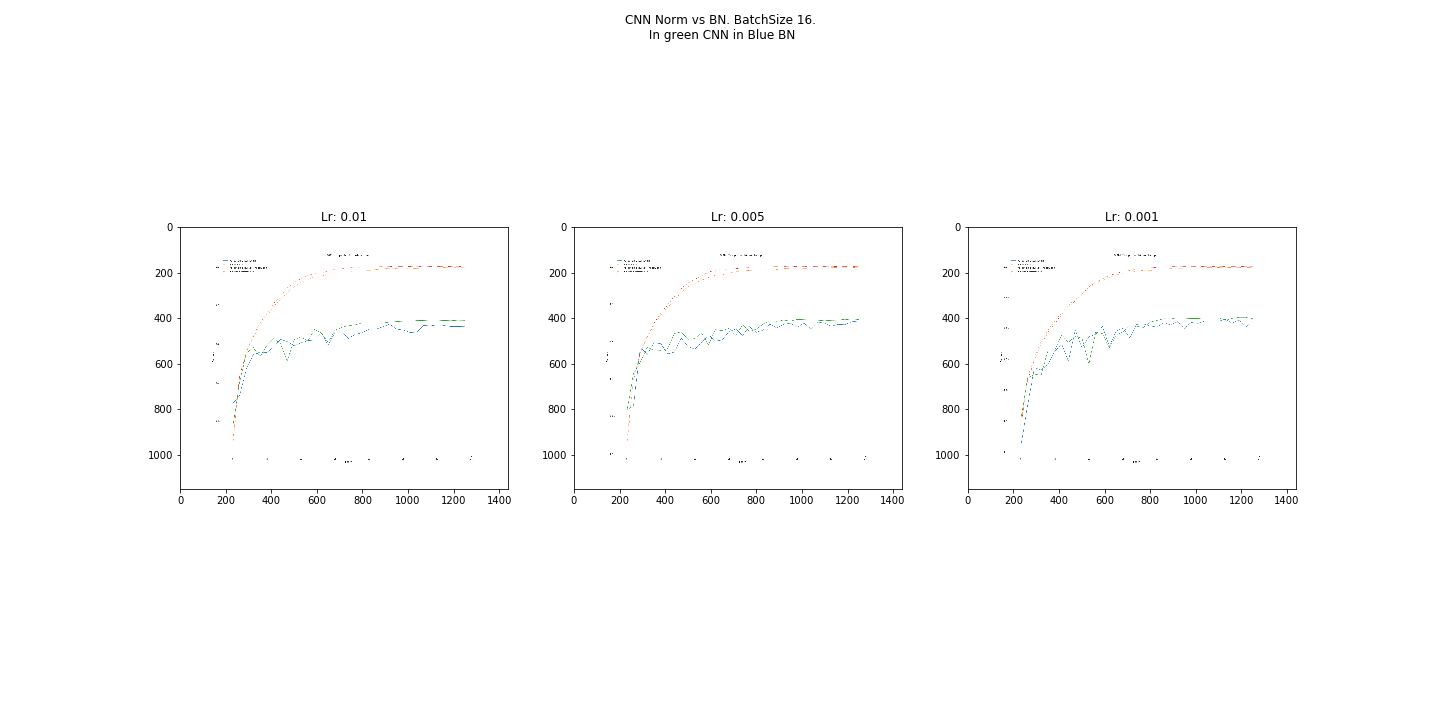}
\end{figure}

\end{document}